\documentclass[10pt,twocolumn,letterpaper]{article}

\usepackage[pagenumbers]{cvpr} 

\usepackage{graphicx}
\usepackage{amsmath}
\usepackage{amssymb}
\usepackage{booktabs}
\usepackage{multirow}
\usepackage{algorithmic}
\usepackage{algorithm}
\usepackage{natbib}  
\usepackage{bibentry}
\usepackage{pifont}
\usepackage[pagebackref=true,breaklinks=true,colorlinks=true,bookmarks=false,linkcolor={red!50!black},urlcolor={darkgray!80!black},citecolor={green!50!black}]{hyperref} 

\usepackage[capitalize]{cleveref}
\crefname{section}{Sec.}{Secs.}
\Crefname{section}{Section}{Sections}
\Crefname{table}{Table}{Tables}
\crefname{table}{Tab.}{Tabs.}

\usepackage{xcolor}
\usepackage{enumitem}
\usepackage{xspace}
\usepackage{booktabs}
\usepackage{colortbl}
\usepackage{multirow}
\usepackage{makecell}
\usepackage{lipsum} 
\usepackage{gensymb} 

\usepackage[labelsep=period]{caption}
\renewcommand{\paragraph}[1]{\vspace{0.2em}\noindent \textbf{#1 \hspace{0.2em}}}

\definecolor{MyDarkRed}{rgb}{0.46, 0.16, 0.16}
\definecolor{MyDarkBlue}{rgb}{0.16, 0.16, 0.66}

\def\paperID{*****} 

\newcommand{\shortname}{FrozenSeg}

\begin{document}

\title{FrozenSeg: Harmonizing Frozen Foundation Models for Open-Vocabulary Segmentation}

\author{%
  Xi Chen \textsuperscript{1} \space \space \space \space
  Haosen Yang \textsuperscript{2} \space \space \space \space
  Sheng Jin \textsuperscript{3} \space \space \space \space
  Xiatian Zhu \textsuperscript{2} \space \space \space \space
  Hongxun Yao \textsuperscript{1}  \space \space \space \space \\
  \textsuperscript{1}Harbin Institute of Technology  \space \space
  \textsuperscript{2} University of Surrey  \space \space
  \textsuperscript{3} Nanyang Technological University \\
}

\maketitle
\begin{abstract}
Open-vocabulary segmentation poses significant challenges, as it requires segmenting and recognizing objects across an open set of categories in unconstrained environments. Building on the success of powerful vision-language (ViL) foundation models, such as CLIP, recent efforts sought to harness their zero-short capabilities to recognize unseen categories. 
Despite notable performance improvements, these models still encounter the critical issue of {\em generating precise mask proposals for unseen categories and scenarios},
resulting in inferior segmentation performance eventually.
To address this challenge, we introduce a novel approach, \textbf{\shortname}, designed to integrate spatial knowledge from a localization foundation model (\eg SAM) and semantic knowledge extracted from a ViL model (\eg CLIP), in a synergistic framework. 
Taking the ViL model's visual encoder as the feature backbone, we inject the space-aware feature into the learnable queries and CLIP features within the transformer decoder.
In addition, we devise a mask proposal ensemble strategy for further improving the recall rate and mask quality.
To fully exploit pre-trained knowledge while minimizing training overhead, we freeze both foundation models, focusing optimization efforts solely on 
a lightweight transformer decoder for mask proposal generation -- the performance bottleneck.
Extensive experiments demonstrate that \textbf{\shortname} advances state-of-the-art
results across various segmentation benchmarks, trained exclusively on COCO panoptic data, and tested in a zero-shot manner. Code is available at \url{https://github.com/chenxi52/FrozenSeg}.
\end{abstract}
\section{Introduction}
Image segmentation is a fundamental task in computer vision, enabling a wide range of applications
such as object recognition~\cite{Mask2Former,SwiftNet}, scene understanding~\cite{P2T,ContextPrior}, and image manipulation~\cite{Generic_Manipulation}. However, traditional techniques are often tailored to specific datasets and segmentation tasks, resulting in a significant gap compared to human visual intelligence, which can perceive diverse visual concepts in the open world. To bridge this disparity, the concept of open-vocabulary segmentation has emerged. In this task, the segmenter is trained to recognize and segment instances and scene elements from any category, mirroring the broad capabilities of human perception.

\begin{figure}[tb]
    \centering
    \includegraphics[width=0.98\linewidth]{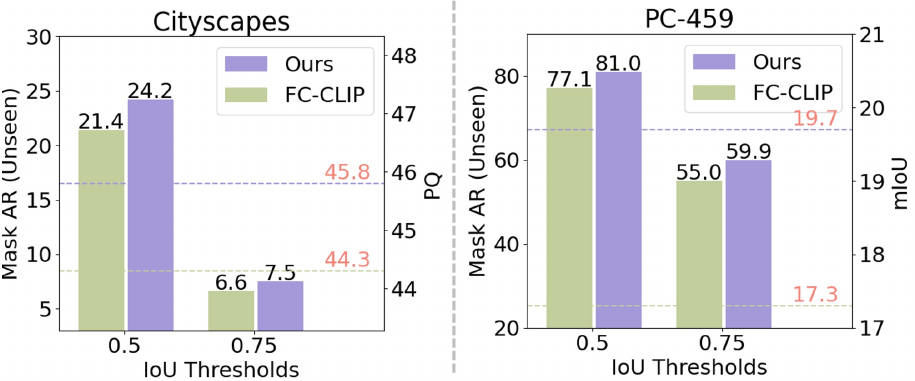}
    \caption{\textbf{Comparision of mask recall of unseen classes and final results performance between FC-CLIP~\cite{fc-clip} and our approach.} Evaluating the performance on the Cityscapes and PC-459 datasets with IoU thresholds of 0.5 and 0.75, our \shortname{} approach significantly increases the mask average recall (AR) of unseen classes and delivers improved final results in Panoptic Quality (PQ) and Mean Intersection-over-Union (mIoU).}
    \label{fig:moti}
\end{figure}

Parallel to these efforts, significant advancements have been made in the field of purpose-generic image-level large-dataset pretrained Vision Language (ViL) representation learning, exemplified by foundational models such as CLIP~\cite{clip} and ALIGN~\cite{ALIGN}. 
These models are pivotal in understanding open scenes, as they leverage rich, descriptive language cues to enhance models’ ability to generalize across a wide array of unseen categories. 
However, the absence of sufficient pixel-level annotations often leads to challenges in dense-level image-text alignment.
Recent studies have utilized these pre-trained ViL models for region classification~\cite{SCAN,SimBase,opensd},
necessitating further training of a segmentation model~\cite{Mask2Former} for precise pixel-level alignment, often resulting in inefficiencies and reduced effectiveness. 
Alternatively, mask proposals generated with the CLIP visual encoder~\cite{fc-clip} are still suboptimal due to their limited fine-grained pixel-level understanding, which becomes a performance bottleneck as the mask proposal generation may overfit to the training classes, undermining the model’s generalizability to unseen classes. As shown in Fig. \ref{fig:moti}, existing methods such as FC-CLIP~\cite{fc-clip} struggle to generalize to unseen categories under different IoU thresholds, significantly limiting their practical utility.

In this paper, to overcome the above limitation, we introduce \shortname{}, a system that harnesses the capabilities of localization foundation model SAM to synergistically and efficiently enhance the coarse semantic features extracted from CLIP by incorporating generalized fine space-aware features.
\shortname{} has three key modules: (1) \texttt{Query Injector}, which aggregate local space-aware features from SAM to serve as the spatial query for the corresponding mask region, enhancing the learnability of queries in a transformer decoder.
(2) \texttt{Feature Injector},
designed to enrich each pixel's CLIP feature by incorporating comprehensive global spatial information from SAM.
(3) \texttt{OpenSeg Ensemble Module}, designed to further boost the quality of mask predictions based on the spatial information injection of SAM during training by ensembling with zero-shot mask proposals from SAM.
Building upon these modules, as shown in Fig.~\ref{fig:moti}, the recall metrics of unseen categories on the challenging CityScapes dataset~\cite{cityscapes} showed significant improvement, consequently boosting PQ from 44.3 to 45.8. This upward trend is further supported by the results in PC-459~\cite{pas_context}, with mIoU increase from 17.3 to 19.7, validating the observed enhancement. 

Our {\bf contributions} can be summarized as follows:
(1) Addressing an acknowledged limitation in mask proposal quality, we introduce \shortname{}, a framework that incorporates foundational models to tackle the open-vocabulary segmentation task effectively.
(2) We propose three critical components: the \texttt{Query Injector}, the \texttt{Feature Injector}, and the \texttt{OpenSeg Ensemble Module}. These components are designed to enhance the integration of SAM features into the transformer decoder, facilitating generalized mask generation.
(3) Extensive experiments on various segmentation tasks demonstrate the superiority of our \shortname{} in generating mask proposals and achieving enhanced final performance, surpassing previous approaches.

\section{Related Works}
\subsection{Open-vocabulary Segmentation}
Open-vocabulary segmentation aims to segment objects even without seeing those classes during training. Previous approaches~\cite{SimBase,OVSeg,SCAN} typically employ a two-stage process, where an additional segmentation model generates class-agnostic mask proposals, which are then interacted with CLIP features. 
In the context of open-vocabulary panoptic segmentation, which necessitates instance segmentation and interaction with multiple mask proposals~\cite{maskclip,opensd}, methods such as OPSNet~\cite{opsnet} combine query embeddings with the last-layer CLIP embeddings and applies an IoU branch to filter out less informative proposals. MaskCLIP~\cite{maskclip} integrates learnable mask tokens with CLIP embeddings and class-agnostic masks.
Despite these advancements, challenges remain in effectively aligning segmenters with CLIP.

Alternatively, one-stage open-vocabulary segmentation faces challenges in extending vision-language models without dedicated segmentation models and addressing overfitting in an end-to-end format. CLIP’s pre-training on image-text pairs necessitates reconciling the region-level biases of the vision-language model. Research such as FC-CLIP and F-VLM~\cite{f-vlm} indicates that convolutional CLIP models generally exhibit superior generalization capabilities compared to ViT-based~\cite{vit} counterparts, primarily due to their capability to handle larger input resolutions effectively. This finding highlights a promising direction for adapting CLIP for improved performance in segmentation tasks.
Despite these advancements, a fundamental issue persists: accurately generating mask proposals for unseen categories and scenarios. This challenge is compounded by the methods' dependence on a static Vision and ViL model, which is not equipped to discern intricate pixel-level details, thereby limiting its effectiveness in mask proposal generation.

\subsection{Large-scale Foundation Models}
Recent advances in large-scale foundation models, pre-trained on extensive datasets, have showcased exceptional zero-shot capabilities. Multi-modal foundation models, such as CLIP and ALIGN, exhibit strong generalization across various downstream tasks. Although these models are trained on image-level data with inherent noise, they can be effectively fine-tuned for various applications. Common strategies include prompt learning~\cite{Coop,DPT} and the use of adapters~\cite{Tip-Adapter}, with CLIP often remaining frozen to preserve its broad generalization.

In the realm of the segmentation foundation models,  significant progress is exemplified by the SAM model~\cite{SAM}, which leverages the extensive SA-1B dataset to achieve notable zero-shot generalization. SAM can adapt to new datasets without additional training by using input prompts. 
Subsequent models, such as HQ-SAM~\cite{hq_sam} and GenSAM~\cite{gensam} have built upon this foundation by optimizing output tokens and integrating textual semantic reasoning, respectively. Despite these advancements, these methods often rely on manually crafted prompts, which constrains their wider applicability and scalability.

Recent research~\cite{Grounded_SAM,regionspot,open-vocabulary_SAM} has explored the use of bounding boxes generated through open-vocabulary detection methods as prompts, combining SAM and CLIP to exploit their complementary strengths in open-vocabulary segmentation. These approaches aim to combine SAM’s zero-shot generalization capabilities with CLIP’s robust feature representations. Despite these efforts, significant challenges remain in achieving fully automatic open-vocabulary segmentation and transitioning from instance segmentation to semantic and panoptic segmentation.

\begin{figure*}[tb]
  \centering
  \includegraphics[width=0.9\linewidth]{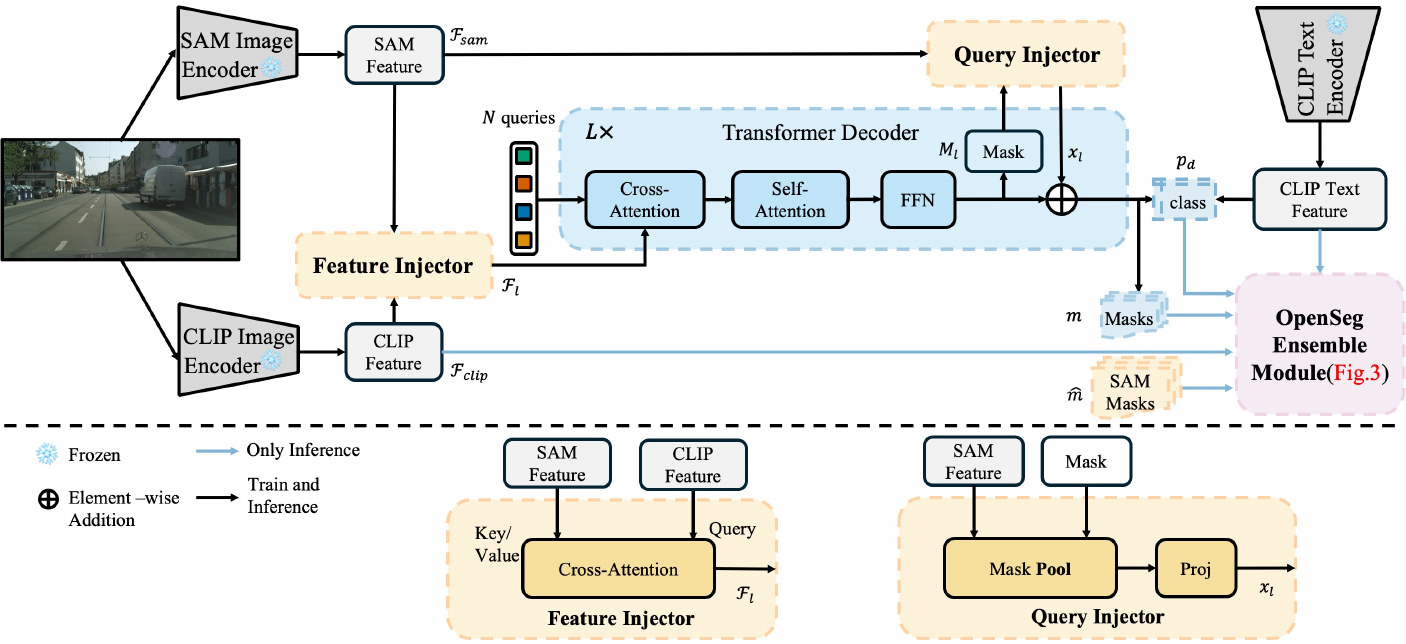}
  \caption{\textbf{Overview of our \shortname{} approach:} (\textbf{Top}) We introduce three key components: the \texttt{Query Injector}, \texttt{Feature Injector} and \texttt{OpenSeg Ensemble Module} to enhance open-vocabulary dense-level understanding. Given $N$ queries, spatial information from SAM is injected into these queries within intermediate layers of the transformer encoder, leading to $N$ class and $N$ corresponding mask predictions. The \texttt{OpenSeg Ensemble Module} then integrates these predictions with zero-shot SAM masks to generate the final results. (\textbf{Bottom}) Detailed design of the two injectors.}
  \label{fig:framework}
\end{figure*}

\begin{figure}[tb]
  \centering
  \includegraphics[width=1.\linewidth]{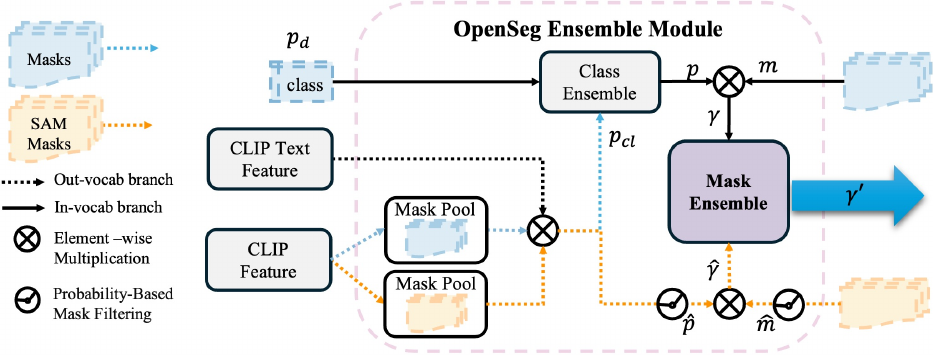}
  \caption{ \textbf{Overview of \texttt{OpenSeg Ensemble Module}}. SAM masks are generated through uniform sampling of point prompts. The module employs a novel \textit{mask ensemble} strategy, injecting SAM mask predictions into unseen mask predictions to enhance the generalization of mask proposals.}
  \label{fig:open_seg}
\end{figure}

\section{Method}
Our objective is to achieve efficient open-vocabulary segmentation using frozen foundation models. In this section, we start by defining the problem. Subsequently, we present our method, \shortname{}, which integrates frozen foundation models for open-vocabulary segmentation through two key components: the \texttt{Query Injector} and the \texttt{Feature Injector}, as illustrated in Fig.~\ref{fig:framework}. Finally, we detail our inference strategy, the \texttt{OpenSeg Ensemble Module}.

\subsection{Problem Definition}
Open-vocabulary segmentation involves training with ground-truth masks corresponding to a predefined set of class labels, $\mathbf{C}_{train}$. 
During testing, the model encounters a different set of class labels, $\mathbf{C}_{test}$, which includes novel classes not seen during training. This process requires segmenting images in an open-world context, where the model must categorize pixels into semantic classes for semantic segmentation, identify individual instances for instance segmentation, or combine both for panoptic segmentation. The notation $\mathbf{C}$ represents either $\mathbf{C}_{train}$ or $\mathbf{C}_{test}$, depending on whether the phase is training or testing.
\label{sec:problem}

\subsection{Our Approach \shortname}
\label{sec:frozenseg}
\paragraph{Overall Architecture}
Following the approach of~\cite{ODISE,fc-clip}, we adopt Mask2Former~\cite{Mask2Former} as our framework. A set of $N$ learnable queries that represent all things and stuff in the input image is processed through the transformer decoder to get mask predictions $m$. 
To adapt the framework for open vocabulary segmentation, we replace the original classification layer with the text embeddings derived from the CLIP text encoder, resulting in class prediction $p_d$, where $d$ denotes the mask detector.
Post-training, the embeddings for each mask and its corresponding category text are projected into a shared embedding space, facilitating effective categorization within the open-vocabulary framework.
In line with~\cite{fc-clip, f-vlm}, we adopt the convolution-based CLIP visual encoder as our image feature extractor, leveraging its pre-trained, frozen weights to obtain high-resolution semantic information.

To address the limitations of CLIP’s coarse features, we introduce two key modules: the~\texttt{Query Injector} and the~\texttt{Feature Injector}. These modules integrate the spatial features from SAM into the mask proposal generation process, as depicted in Fig. \ref{fig:framework}.
Unlike ~\cite{vit-adapter}, which incorporates multi-level spatial information into the vision transformer, our injectors focus on infusing spatial information directly into mask queries.
Additionally, we propose the \texttt{OpenSeg Ensemble Module} to further enhance segmentation performance during inference.
We detail our approach below: 

\paragraph{Query Injector} To improve local spatial understanding, we introduce the Query Injector, which enhances the learnable query with space-aware features derived from SAM. 
The transformer decoder uses masked multi-head attention to bolster cross-attention between the image's foreground region and the learnable queries. 
This mechanism facilitates the integration of both content and spatial information within the query, a concept supported by prior studies such as~\cite{detr,dab-detr,conditional_Detr}.
However, capturing detailed spatial information remains challenging when the backbone is frozen.

To address this challenge, we devise the Query Injector, which leverages newly generated masks at each decoder layer to pool and transform SAM visual features into a spatial query. The process for generating the spatial query is defined as follows:
\begin{equation}
x_l = f(\textbf{\text{pool}}(M_l, \mathcal{F}_{sam}))
\end{equation}

Here, $l$ represents the layer index in the transformer decoder, $f$ denotes a linear projection function, and \textbf{pool} refers to the mask pooling operation.  $\mathcal{F}_{sam}$ represents the SAM-derived image features. 
This spatial query is specifically designed to concentrate on a region encompassing the mask region. Subsequently, the spatial query is integrated with the learnable query via element-wise addition.

\paragraph{Feature Injector}
To refine the CLIP features for mask generation on a global scale, we introduce the Feature Injector, which uses the multi-head cross-attention mechanism (MHCA) as detailed in \cite{transformer}. This mechanism is renowned for its effectiveness in amalgamating diverse information. In our approach, we extend MHCA to enhance the coarse semantic features from CLIP.
Specifically, the Feature Injector integrates semantic content from CLIP with pixel-level spatial awareness from SAM, providing a more nuanced understanding at the pixel scale.
The mathematical formulation of this feature integration is as follows:
\begin{equation}
\mathcal{F} = \text{SoftMax} \left(\frac{ f_q (\mathcal{F}_{clip}) \cdot f_k( \mathcal{F}_{sam})}{\sqrt{D}} \right) \cdot f_v(\mathcal{F}_{sam})
\end{equation}

Here, $f_q$, $f_k$, and $f_v$ are linear projection functions in MHCA. $\mathcal{F}_{clip}$ and $\mathcal{F}_{sam}$ represent the features extracted from CLIP and SAM, respectively, while $D$ denotes the dimensionality of the projected features.

\paragraph{Inference Strategy}
\label{sec:train_infer}
Previous works such as~\cite{fc-clip,f-vlm,freeseg} have validated the efficacy of mask pooling on CLIP features within class ensemble methodologies to enhance open-vocabulary segmentation capabilities. Building on these techniques, our approach introduces a novel mask ensemble strategy. As illustrated in Fig.~\ref{fig:open_seg}, our OpenSeg Ensemble Module initiates with the class ensemble process:
\begin{equation}
p_i(j) =
    \begin{cases}
     (p_{i,d}(j))^{(1-\alpha)} \cdot (p_{i,cl}(j))^{\alpha}, &\text{if $j\in \mathbf{C}_{train}$ } \\
    (p_{i,d}(j))^{(1-\beta)} \cdot (p_{i,cl}(j))^{\beta}, &\text{else}
    \end{cases}
\label{eq:class_ensemble}
\end{equation}

Here, $p_i(j)$ denotes the combined probability for class $j$ in proposal $i$, integrating inputs from both the detector ($p_{i,d}$) and CLIP ($p_{i,cl}$).
The mask predictions $r$ for $N$ queries are then generated by aggregating the products of these probability-mask pairs:
$ \sum_{i=1}^N p_i(c)\cdot  m_i[x,y] = r \in \mathbb R^{C\times HW}$.

Drawing inspiration from the class ensemble, we utilize zero-shot mask predictions from SAM to perform a mask ensemble on $r$. The SAM masks, denoted as $M_{sam}=\{\hat m_i\}_{i=1}^N$, are generated by uniformly sampling points prompts across the image. 
These masks are used to pool CLIP features and derive classification scores $P_{sam} = \{\hat p_i\}_{i=1}^N$ by aligning with CLIP text features. A threshold $\xi=0.5$ is applied to filter these masks based on the maximum probability, resulting in the selected probability-mask pairs $\{(\hat{m}_i,\hat {p}_i) \mid \operatorname{argmax}_c\hat{p}_i > \xi\}_{i=1}^{N'}$.

In the context of semantic segmentation, the SAM mask predictions, denoted as $\hat{r}$, are computed similarly as follows: $\hat r = \sum_{i=1}^{N'} \hat p_i(c)\cdot \hat m_i[x,y] $. The final mask prediction, $r^{'}$, is obtained by integrating the predictions $r$ and $\hat{r}$ through a mask ensemble approach:
\begin{equation}
r'[x,y](j) =
    \begin{cases}
     r[x,y](j), \text{if $j\in \mathbf{C}_{train}$ } \\
    (1-\epsilon)*r[x,y](j) + \epsilon * \hat r[x,y](j), \text{else}
    \end{cases}
\label{eq:ensemble}
\end{equation}

Subsequently, the final semantic segmentation results are determined by assigning each pixel $[x,y]$ a class based on 
$\operatorname{argmax}_{c\in\{1,...,|\mathbf{C}|\}} r'[x,y]$.

In the context of panoptic segmentation, the efficacy of the results significantly depends on the performance of individual queries. This dependency reduces the effectiveness of integrating class-agnostic mask predictions. Therefore, the final results are determined by assigning each pixel to one of the $N$ predicted probability pairs. This assignment is performed through the following expression: $\operatorname{argmax}_{i:c_i \neq \emptyset}p_i(c_i)\cdot m_i[x,y]$. In this expression, $c_i$ represents the most likely class label, which is determined by $c_i=\operatorname{argmax}_{c\in\{1,...,|\mathbf{C}|,\emptyset\}} p_i(c)$. Here, $\emptyset$ denotes the class of 'no object'.

\section{Experiments}
\begin{table*}[tb]\centering
    \caption{\textbf{Performance of open-vocabulary panoptic segmentation. 
    } We present results obtained using both CLIP RN50x64 and ConvNext-L. \textbf{Bold} represents best, \underline{underline} indicates second best. * denotes re-implemented final results.}
    \label{tab:panoptic_seg_final}
    \setlength{\tabcolsep}{1mm}
    \small
    \begin{tabular}{l|c|ccc|ccc|cc|cc|ccc}
        \toprule
        & & \multicolumn{3}{c|}{ADE20K} &  \multicolumn{3}{c|}{Cityscapes} & \multicolumn{2}{c|}{Mapillary Vistas} &\multicolumn{2}{c|}{BDD100K}&\multicolumn{3}{c}{COCO (seen)}\\
       Method & ViL Model & PQ & AP & mIoU & PQ & AP & mIoU & PQ & mIoU &PQ&mIoU& PQ & AP & mIoU\\
        \midrule
        \hline
        OPSNet&RN50&19.0 &-&25.4&41.5&-&-&-&-&-&-&\textbf{57.9}&-&64.8\\
        \midrule
        MaskCLIP&ViT-L/14 &15.1 &6.0& 23.7&-&-&-&-&-&-&-&30.9&-&-\\
        MaskQCLIP&ViT-L/14 &23.3&-&30.4&-&-&-&-&-&-&-&-&-&-\\
        ODISE&ViT-L/14&23.4 &\underline{13.9}& 28.7&-&-&-&-&-&-&-&45.6& 38.4 &52.4\\
        ODISE+CLIPSelf&ViT-L/14&23.7 &13.6& 30.1&-&-&-&-&-&-&-&45.7& 38.5& 52.3\\
        
        \midrule
        FC-CLIP*& RN50x64 & 21.3& 13.2 &28.7 &42.6 &27.3 &55.1 &\underline{18.2}&27.4&13.8& 41.4&55.3&46.5&64.8\\
        \textbf{\shortname} & RN50x64 &23.1&13.5&30.7&\underline{45.2}&\textbf{28.9}&\underline{56.0}&18.1 &\underline{27.7}&12.9&46.2&55.7& \textbf{47.4} &\underline{65.4} \\
        \midrule
        
        FC-CLIP*& ConvNeXt-L & \underline{25.1} &\textbf{16.4}& \underline{32.8} & 44.3& 27.9 &\underline{56.0} &18.1 &\textbf{27.9}&\underline{17.9}&\underline{49.4}&\underline{56.4}&\textbf{47.4}&65.3\\
        \textbf{\shortname} & ConvNeXt-L & \textbf{25.9}&\textbf{16.4}&\textbf{34.4}&\textbf{45.8}&\underline{28.4}&\textbf{56.8}&\textbf{18.5}&27.3&\textbf{19.3}&\textbf{52.3}&56.2& \underline{47.3} &\textbf{65.5} \\
        \bottomrule
    \end{tabular}
\end{table*}

\subsection{Datasets and Evaluation Protocal}
For training, we use the COCO panoptic~\cite{COCO} dataset, which includes 133 classes. Our evaluation covers open-vocabulary panoptic, semantic, and instance segmentation tasks in a zero-shot setting spanning several test datasets. In semantic segmentation, we access performance on ADE20K dataset~\cite{ADE}, which includes both a subset with 150 classes (A-150) and a full version with 847 classes (A-847). Additionally, we evaluate on PASCAL VOC~\cite{pasvoc}(PAS-21), which has 20 object classes and one background class, and PASCAL-Context~\cite{pas_context}, an extension of PASCAL VOC with 459 classes (PC-459). For panoptic segmentation, the datasets used are ADE20K~\cite{ADE}, Cityscapes~\cite{cityscapes}, Mapillary Vistas~\cite{mapillary}, and BDD100K~\cite{bdd100k}, alongside the closed-set COCO validation dataset. 
For instance segmentation, we choose to evaluate LVIS v1.0~\cite{lvis}, which features 337 rare categories.
The evaluation metrics include mean intersection-over-union (mIoU) and frequency-weighted IoU (FWIoU) that offer a comprehensive evaluation of overall performance for semantic segmentation, panoptic quality (PQ), average precision (AP), and mIoU for panoptic segmentation, as well as AP for instance segmentation.

\subsection{Baselines} 
We compare with multiple state-of-art approaches as follows: OPSNet~\cite{opsnet},
MaskCLIP~\cite{maskclip}, MasQCLIP~\cite{MasQCLIP}, ODISE~\cite{ODISE}, CLIPSelf~\cite{clipself}, FC-CLIP~\cite{fc-clip}, Ovseg~\cite{OVSeg}, SAN~\cite{san} , RegionSpot~\cite{regionspot} and Open-Vocabulary SAM~\cite{open-vocabulary_SAM}.

\subsection{Implement Details}
\label{sec:implement}
We use 250 queries for both training and testing, with CLIP serving as the backbone for open-vocabulary text-image alignment. Specifically, we employ the RN50x64 and ConvNext-Large~\cite{ConvNext} versions of CLIP. Additionally, we validate our approach using the ViT-Base~\cite{vit} model from SAM, with the selection rationale detailed in the \texttt{Supplementary}.
To obtain multi-level semantic features, we apply feature pyramid networks (FPN) after CLIP. SAM processes input images $\mathbf{I} \in \mathbb{R}^{H, W}$, where $H = W = 1024$. As demonstrated by PlainViT~\cite{plainVit}, the deepest feature of ViT contains sufficient information for multi-scale object recognition, and given that SAM is frozen, we do not use FPN for SAM. Instead, we utilize a single convolution layer to project the features to the necessary resolution and then feed them into a single-scale deformable attention transformer~\cite{DEFORMABLE_DETR} as the pixel decoder in the SAM branch.
Our transformer decoder comprises $L=9$ layers. Feature maps with resolutions of 1/8, 1/16, and 1/32 are processed by successive decoder layers in a round-robin fashion. During training, we follow the strategy and losses outlined in FC-CLIP, selecting the model from the final iteration for our primary results. Training is conducted on 4 Tesla A100 GPUs with a batch size of 16.

\subsection{Inference Details}
During inference, we adhere to the FC-CLIP by resizing images such that the shortest side is 800 pixels for general datasets and 1024 for the Cityscapes and Mapillary Vistas datasets.
We employ a 32x32 grid of points research to generate masks from the SAM ViT-Huge model. The default parameters are set as follows: $\alpha=0.4$ and $\beta=0.8$ in Eq.\eqref{eq:class_ensemble}, and a mask ensemble parameter $\epsilon=0.2$ in Eq.~\eqref{eq:ensemble}.

\begin{figure}[t]
    \centering
    \includegraphics[width=0.98\columnwidth]{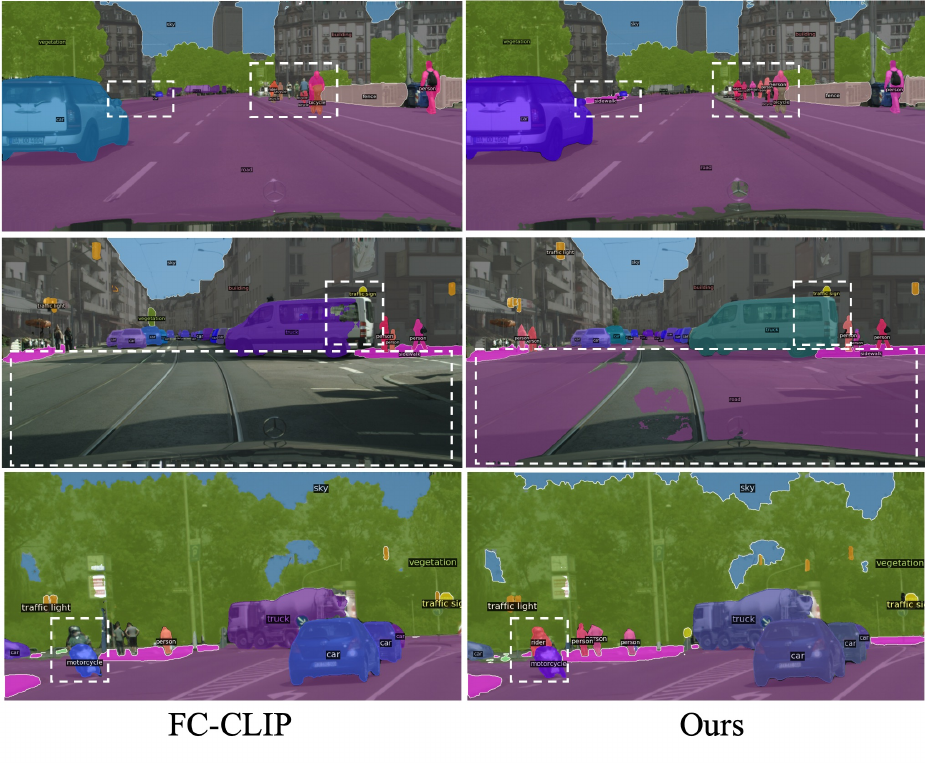}
    \caption{\textbf{Qualitative illustration of panoptic segmentation results} on Cityscapes. White boxes highlight areas with notable differences between methods. Compared to FC-CLIP, \shortname{} shows improved performance in predicting small objects (row 1), more accurate entity segmentation (row 2), and better generalization to the unseen class 'rider' (row 3). }
    \label{fig:vis_panop}
\end{figure}
\begin{table*}[tb]\centering
    \caption{\textbf{Performance of cross-dataset open-vocabulary semantic segmentation.} 'IN' refers to the ImageNet(50K)~\cite{ImageNet} dataset, and 'Panop.+Cap.' signifies the combined use of COCO panoptic~\cite{COCO} and COCO caption~\cite{coco_caption} datasets. \textbf{Bold} represents best, \underline{underline} indicates second best. * denotes the results from re-implemented final evaluations. } 
    \label{tab:semantic_seg_final}
    \small
    \setlength{\tabcolsep}{1mm}
        
        
        
    
    \begin{tabular}{l|c|c|cc|cc|cc|cc}
        \toprule
       & & &  \multicolumn{2}{c|}{PC-459} &\multicolumn{2}{c|}{PAS-21} &\multicolumn{2}{c|}{A-847}&\multicolumn{2}{c}{A-150} \\
       Method& ViL Model &Training Dataset&mIoU&FWIoU&mIoU&FWIoU&mIoU&FWIoU&mIoU&FWIoU\\
        \midrule
        \hline
        OPSNet&RN50&COCO Panop.+IN&-&-&-&-&-&-&25.4&-\\
        \midrule
        
        Ovseg &ViT-L/14&COCO Stuff &12.4&- &-&-&9.0&-&29.6&-\\
        SAN &ViT-L/14&COCO Stuff &17.1&- &-&-&13.7&-&\underline{33.3}&- \\
        MaskCLIP &ViT-L/14&COCO Panoptic&10.0&-& -&-&8.2&-&23.7\\
        MasQCLIP &ViT-L/14&COCO Panoptic &18.2&-&-&-&10.7&-&30.4\\
        ODISE&ViT-L/14 &COCO Panop.+Cap. &13.8&-&\underline{82.7}&-&11.0&-&29.9\\
        ODISE+CLIPSelf&ViT-L/14& +COCO Stuff &-&-&-&-&-&-&30.1&-\\

        \midrule
        FC-CLIP* &RN50x64&COCO Panoptic&15.6&50.0&81.1&91.4&10.8&44.6&28.7&52.5\\
        \textbf{\shortname} & RN50x64 &COCO Panoptic& \underline{18.7}&\underline{60.1}&82.3&\underline{92.1}&11.8&\textbf{52.8}&30.7&\underline{56.6}\\
        \midrule
        
        FC-CLIP*&ConvNeXt-L&COCO Panoptic &17.3 &56.7& \textbf{83.0}&\textbf{92.4} &\underline{14.0}& 48.1&32.8&56.1\\
        \textbf{\shortname}&ConvNeXt-L& COCO Panoptic&\textbf{19.7} &\textbf{60.2}&82.5&\underline{92.1}&\textbf{14.8}&\underline{51.4}&\textbf{34.4}&\textbf{59.9}\\
        \bottomrule
    \end{tabular}
    
\end{table*}
\begin{figure*}[!thb]
  \centering
  \includegraphics[width=0.8\textwidth]{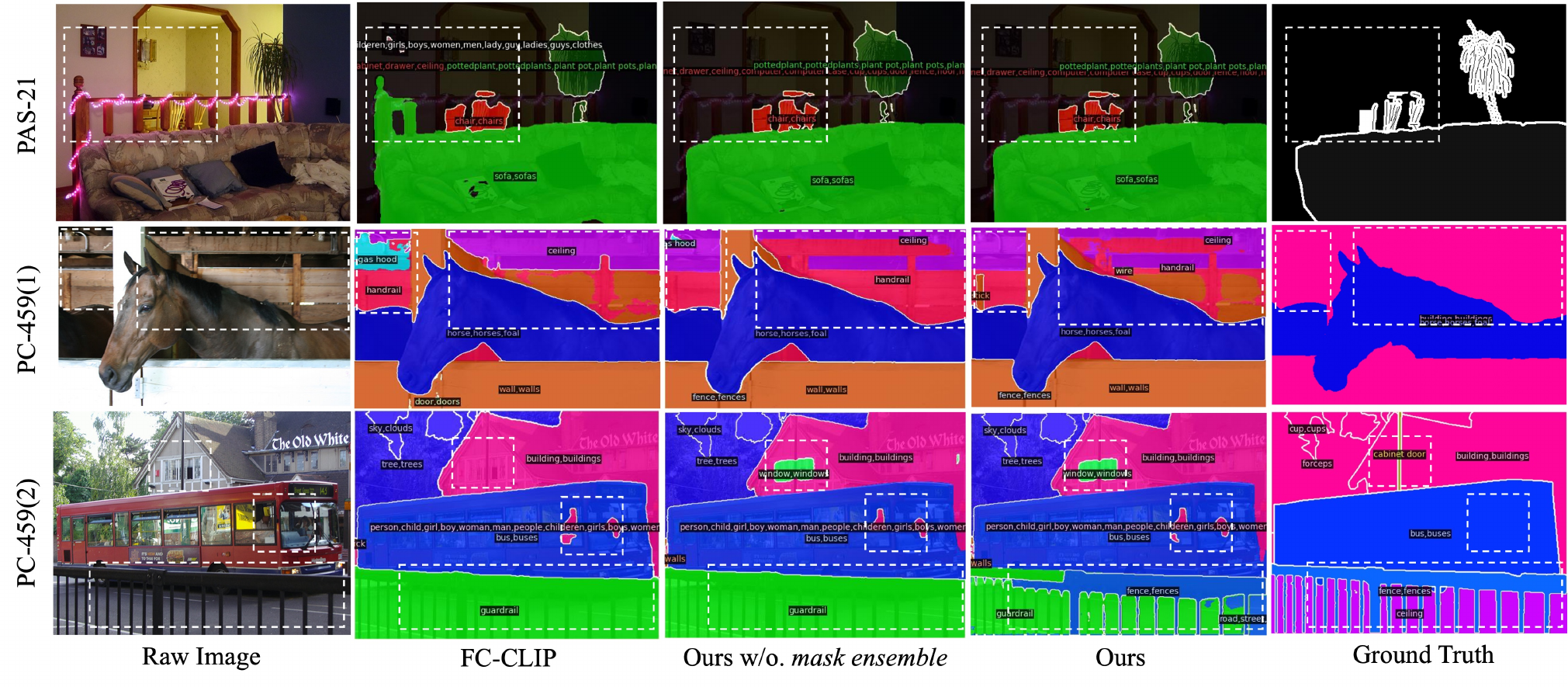}
  \caption{\textbf{Qualitative comparison of semantic segmentation results}. White boxes indicate areas of discrepancy. Our \shortname{} (col. 4) has contextually appropriate results compared to FC-CLIP (col. 2) and ground truth annotations (col. 5). 
  }
  \label{fig:vis_seg}
\end{figure*}

\begin{table}[!tb]\centering
    \caption{Performance ($\text{AP}_r$) of open-vocabulary instance segmentation on rare categories in LVIS v1.0 dataset. * denotes the results from re-implemented final results.} 
    \label{tab:lvis}
    \small
    \setlength{\tabcolsep}{1mm}
    \begin{tabular}{l|c|c}
        \toprule
       Method& Proposals &$\text{LVIS}_r$ \\ 
        \midrule
        \hline 
        \multirow{3}{*}{RegionSpot}&GLIP-T(B)~\cite{GLIP}&12.7\\
        &SAM&14.3\\
        &GLIP-T~\cite{GLIP}&20.0\\
        \cmidrule(lr){1-3}
        Open-Vocabulary SAM&Detic~\cite{Detic}&24.0	\\
        FC-CLIP*&-&25.0\\
        \midrule
        \textbf{\shortname} &-&\textbf{25.6}\\
        \bottomrule
    \end{tabular}
\end{table}
\begin{table*}[!thb]\centering
    \caption{\textbf{Ablations of the proposed modules}: results following complete training iterations with mIoU metrics. 'Inj.': Injector.}
    \label{tab:abl_method}
    \setlength{\tabcolsep}{1mm}
    \small
    \begin{tabular}{l|ccc|cc|*3{c}}
        \toprule
       &&&&PC-459&A-847&\multicolumn{3}{c}{ADE20K} \\
    \multicolumn{1}{c|}{\#} &\texttt{Query Inj.} & \texttt{Feature Inj.} & \texttt{OpenSeg Ensemble} &mIoU&mIoU&PQ&AP&mIoU\\
        \midrule
        \midrule
       (1) only SAM & \ding{55} & \ding{55} & \ding{55}&  6.6&6.5&-&-&25.4\\
       (2) Baseline &\ding{55} & \ding{55} & \ding{55}& 17.3& 14.0&25.1&16.4&32.8\\
       (3) - &\ding{55} & \ding{55}&\checkmark&17.6 &14.5&-&-&33.5\\
       (4) - &\checkmark & \ding{55} & \ding{55}& 18.1&14.2&25.7&16.5&33.6 \\
       (5) - & \checkmark & \checkmark & \ding{55} &18.5 &14.4 &25.9&16.5&33.8\\
      (6) \shortname &\checkmark & \checkmark & \checkmark& \textbf{19.7} &\textbf{14.8}&-&-&\textbf{34.4}\\
        \bottomrule
    \end{tabular}


\end{table*}
\begin{table*}[!thb] \centering
    \begin{minipage}[t]{0.48\linewidth}
        \caption{\textbf{Impact of selected insertion layer in transformer decoder on \texttt{Query Injector}} performance: results after 55K iterations. The 'Size' column is the relative interacted image feature size of multi-level feature maps. }
        \label{tab:abl_query}
        \setlength{\tabcolsep}{1mm}
        \small
        \centering
        \begin{tabular}{l|c|ccc|ccc}
            \toprule
            & &\multicolumn{3}{c|}{COCO} & \multicolumn{3}{c}{Cityscapes}\\
            Size & Layers & PQ &AP & mIoU & PQ & AP & mIoU\\
            \midrule
            \hline
            1/32 & 1, 4, 7 &52.6&42.7&62.6 &\textbf{40.4}&20.8&53.0\\
            1/16 & 2, 5, 8  &52.5&42.5&62.4 &40.1&20.7&53.3 \\
            1/8 & 3, 6, 9 & \textbf{52.7}&\textbf{42.8} &\textbf{62.7} & 40.0&\textbf{21.6}&\textbf{54.0} \\
            \bottomrule
        \end{tabular}
    \end{minipage}\hfill
    \begin{minipage}[t]{0.48\linewidth}
        \caption{\textbf{Comparative analysis of FPS performance and Trainable vs. Frozen parameter counts using a single A100.} All results are obtained from the average time on the validation set, including post-processing.}
        \label{tab:fps}
        \setlength{\tabcolsep}{1mm}
        \small
        \centering
        \begin{tabular}{l|cc|cc|cc}
            \toprule
             & \multicolumn{2}{c|}{COCO} & \multicolumn{2}{c|}{Cityscapes}&\multicolumn{2}{c}{Parmas[M]}\\
            Method & PQ & FPS & PQ & FPS & Frozen & Trainable\\
            \midrule
            \midrule
            FC-CLIP& 56.4 &2.71&44.3& 0.87&200.0&21.0\\
            \shortname & 56.2 &2.15&45.8 &0.78&293.5&26.5\\
            \bottomrule
        \end{tabular}
    \end{minipage}


\end{table*}

\subsection{Evaluation on Open-vocabulary Segmentation}

\subsubsection{Open-vocabulary Panoptic Segmentation}
Tab.~\ref{tab:panoptic_seg_final} presents a comparison of \shortname{} with leading methods in zero-shot open-vocabulary panoptic segmentation. Our approach, \shortname{} with RN50x64, notably surpasses other works and the baseline FC-CLIP, achieving improvements of +1.8 PQ, +0.3 AP, and +2.0 mIoU on ADE20K; +2.6 PQ, +1.6 AP, and +0.9 mIoU on Cityscapes; and +4.8 mIoU on BDD100K.
Additionally, the \shortname{} configuration with ConvNeXt-L delivers enhanced performance on open-set datasets without compromising results on the closed-set COCO validation dataset. Significant improvements include +0.8 PQ and +1.6 mIoU on ADE20K, +1.5 PQ, +0.5 AP, and +0.8 mIoU on Cityscapes, +0.4 PQ on Mapillary Vistas, and +1.4 PQ and +2.9 mIoU on BDD10K. 
Qualitative results of panoptic segmentation on Cityscapes, depicted in Fig.~\ref{fig:vis_panop}, show improvements in segmentation, particularly for small objects, entity recognition, and novel class recognition. Additional details and additional results are available in the \texttt{Supplementary}.

\subsubsection{Open-vocabulary Semantic Segmentation}
Tab.~\ref{tab:semantic_seg_final} presents a comparative analysis of \shortname{} in cross-dataset open-vocabulary semantic segmentation. Using the RN50x64 backbone, \shortname{} significantly outperforms the baselines. Compared to FC-CLIP, \shortname{} achieves gains of +3.1 mIoU and +10.1 FWIoU on PC-459, +1.0 mIoU and +8.2 FWIoU on A-847, and +2.0 mIoU and +4.1 FWIoU on A-150. These improvements are also reflected in the ConvNeXt-L configuration. Overall, \shortname{} sets a new benchmark in performance across the datasets PC-459, A-847, and A-150. It is important to note that PAS-21’s categories fully overlap with those of the training dataset, which suggests that FC-CLIP may overfit to base classes and thus limit its generalization.
For qualitative insights, refer to Fig.~\ref{fig:vis_seg}, where \shortname{} delivers segmentations that are contextually more accurate compared to both the baseline and ground truth annotations, demonstrating its effective handling of complex scenes. Additional results are available in the \texttt{Supplementary}.

\subsubsection{Open-vocabulary Instance Segmentation}
Tab.~\ref{tab:lvis} presents results for rare categories in the LVIS dataset. We compare \shortname{} with approaches that integrate SAM with CLIP for open-vocabulary segmentation tasks, specifically RegionSpot and Open-Vocabulary SAM. Both of these methods rely on proposals as location prompts and are trained on datasets beyond the COCO panoptic to align the models. Our method achieves the highest performance, with an improvement of +0.6 AP over FC-CLIP.

\subsection{Ablation Studies}
We perform a series of ablation studies on our method. All findings are presented using the ConvNext-L version of CLIP and the ViT-B version of SAM.

\subsubsection{The Effectiveness of Each Component}
We perform ablation studies to assess the effectiveness of each component of our method. Tab.~\ref{tab:abl_method} presents the results of these ablations on three challenging out-of-vocabulary datasets, with rows 2-6 highlighting the contribution of each component to overall performance.
Specifically, row 1 illustrates the scenario where only proposals from SAM are utilized. In this setup, SAM masks are used to pool CLIP features, providing basic semantic understanding without explicit semantic guidance. This configuration achieves approximately 6.5 mIoU on PC-459 and A-847, and 25.4 mIoU on ADE20K, demonstrating the fundamental generalization capability of SAM masks.
Therefore, we integrate \texttt{OpenSeg Ensemble} to address the limitation of unseen mask proposals. This enhancement is evident in the comparison between ablation cases (2) and (3), and also between cases (5) and (6).  Fig.~\ref{fig:vis_seg} provides a clearer visual comparison, showing improved segmentation accuracy for objects such as ‘people’ and ‘fences’ in the PC-459(2) example, particularly in columns 3 and 4.

\subsubsection{Where to Inject}
Tab.~\ref{tab:abl_query} presents an ablation examining the impact of layer insertion for the \texttt{Query Injector} within the transformer decoder, which consists of a total of 9 layers. Since SAM Vision Transformers provide the final layer features as the most relevant feature maps, we explore the optimal layer for query injection based on its interaction with corresponding CLIP feature maps. The results indicate that injecting SAM query features at layers $l=3, 6, 9$ yields the most significant improvement, with layer $l+1$ leveraging the newly introduced queries for further refinement.
For the \texttt{Feature Injector}, due to the exponential increase in computational complexity associated with cross-attention computations as feature size expands, we restrict the application of the Feature Injector to 1/32-sized features, specifically at layers $l=1, 4, 7$.

\subsubsection{Speed and Model Size}
As shown in Tab.~\ref{tab:fps}, incorporating SAM along with two custom injectors results in a slight reduction in inference speed, manifesting as a 0.56 and 0.09 decrement in frames per second (FPS) during single-image processing. Despite this, the adjustment leads to a notable improvement of 1.8 PQ on the Cityscapes datasets, with minimal impact on COCO. This reflects a well-balanced trade-off between enhanced performance and computational efficiency. Compared to FC-CLIP, our model requires a modest increase of only 5.5M training parameters and 93.5M frozen parameters, demonstrating its effectiveness.

\section{Conclusion}
In this study, we introduced \textbf{\shortname{}}, a method designed to enhance mask proposal quality in open-vocabulary segmentation by leveraging SAM’s dense-prediction capabilities. \shortname{} employs the \texttt{Query Injector} and \texttt{Feature Injector} modules to integrate SAM visual features with learned queries and CLIP visual features, thereby refining mask proposals through multiple transformer decoder layers. Additionally, we introduce the \texttt{OpenSeg Ensemble Module} for inference, which aggregates zero-shot SAM masks to improve out-vocabulary predictions further. Our experiments demonstrated that \shortname{} significantly enhances the mask proposal quality in open-vocabulary scenarios, highlighting its versatility.

\begin{table*}[!thb] 
\centering
    \caption{Comparative analysis of FC-CLIP models' performance on ADE20K, Cityscapes~\cite{cityscapes}, COCO, PC-459, and A-847. Results in the first row are sourced from the paper~\cite{fc-clip} and are based on checkpoints selected from the ADE20K validation set. The * denotes outcomes from the \textbf{final iteration}. FC-CLIP* is re-implemented using the official code.}
\label{tab:fc-clip}
    \small
    \setlength{\tabcolsep}{1mm}
    \begin{tabular}{l|ccc|ccc|ccc|cc|cc}
        \toprule
        &\multicolumn{3}{c|}{ADE20K}&\multicolumn{3}{c|}{Cityscapes}&\multicolumn{3}{c|}{COCO(seen)}&\multicolumn{2}{c|}{PC-459}&\multicolumn{2}{c}{A-847} \\ 
         Method & PQ & AP & mIoU &PQ & AP & mIoU & PQ & AP & mIoU &mIoU &FWIoU &mIoU&FwIoU \\
        \midrule
        \midrule
        FC-CLIP~\cite{fc-clip} & \textbf{26.8} & \textbf{16.8}&34.1 & 44.0&26.8&56.2&54.4& 44.6 &63.7&18.2& 58.2&\textbf{14.8}&51.3\\
         FC-CLIP* &25.1&16.4&32.8&44.3 & 27.9&56.0&\textbf{56.4}& \textbf{47.4} &65.3&17.3&56.3 &14.0& 48.1 \\
         \midrule
          Ours(*)&25.9&16.4& \textbf{34.4}&\textbf{45.8} &\textbf{28.4} &\textbf{56.8}&56.2& 47.3& \textbf{65.5}&\textbf{19.7}& \textbf{60.2}&\textbf{14.8}&\textbf{51.4}\\
          \bottomrule
    \end{tabular}



\end{table*}
\begin{table*}[!htb]
    \caption{Evaluating open-vocabulary recall with semantic segmentation annotations on Cityscapes, PC-459, A-847, and PAS-21~\cite{pasvoc}: insights into seen (S) and unseen (U) Classes.}
    \setlength{\tabcolsep}{1mm}
    \small
    

    \centering
    \label{tab:recall_all}
    \begin{tabular}{l|ccc|ccc|ccc|ccc}
        \toprule
        & \multicolumn{3}{c|}{Cityscapes(S/U)} & \multicolumn{3}{c|}{PC-459(S/U)} & \multicolumn{3}{c|}{A-847(S/U)}& \multicolumn{3}{c}{PAS-21(S)}\\
        Method & 0.5 & 0.75 & 0.9 & 0.5 & 0.75 & 0.9 &0.5 & 0.75 & 0.9 &0.5 & 0.75 & 0.9  \\
        \midrule
        \midrule
        FC-CLIP&\textbf{68.5}/21.4 &45.6/6.6 &10.3/0.0 &91.2/77.1 &73.1/55.0&43.3/27.7&\textbf{84.2}/65.8 & \textbf{59.6}/40.3& \textbf{27.8}/13.8&90.6&75.6&\textbf{49.0}\\
         Ours &68.2/\textbf{24.2} &\textbf{46.0/7.5}&\textbf{13.3/0.1} 
         &\textbf{92.6/81.0} &\textbf{76.0/59.9}&\textbf{47.0/31.4}
         &78.9/\textbf{82.9}& 52.3/\textbf{59.4}& 23.6/\textbf{24.6}
         &\textbf{92.0}&\textbf{76.4}&46.1\\
    \bottomrule
    \end{tabular}

\end{table*}
\begin{table*}[!thb]\centering
    \caption{Open-vocabulary performance with SAMs on Cityscapes, Mapillary Vistas~\cite{mapillary}, PC-459, A-847, and PC-59~\cite{pas_context}. \textbf{Bold} highlights optimal results.}
    \label{tab:sam_all}
    \centering
    \small
    \setlength{\tabcolsep}{1mm}
    \begin{tabular}{l|ccc|cc|cc|cc|cc}
        \toprule
          & \multicolumn{3}{c|}{Cityscapes} & \multicolumn{2}{c|}{Mapillary Vistas}& \multicolumn{2}{c|}{PC-459}& \multicolumn{2}{c|}{A-847} &\multicolumn{2}{c}{PC-59}\\
         SAM & PQ &  AP & mIoU & PQ & mIoU &mIoU & FWIoU &mIoU & FWIoU&mIoU & FWIoU\\
        \midrule
        \midrule
         ViT-T~\cite{mobile_sam} & 43.0 &25.9 & 55.0 &\textbf{18.6} & \textbf{28.0} &17.6&57.0 &14.0 &48.1&  56.8&66.1 \\
         ViT-B~\cite{SAM} &\textbf{45.8} &\textbf{28.4}&\textbf{56.8} &18.5&27.3 &\textbf{18.5}&\textbf{59.0} &\textbf{14.4} &\textbf{51.5}&  \textbf{58.1}&\textbf{68.7}\\
         ViT-L~\cite{SAM} & 44.2 &27.1&56.2&18.5&27.5&17.4&57.0 &14.2 &48.6&  56.7&66.2\\
         ViT-H~\cite{SAM} & 44.2 & 28.0& 56.2 &18.4 &27.4 &17.3&56.7 &13.9 &48.1&56.6&66.2\\
        \bottomrule
    \end{tabular}
    

    
\end{table*}

\begin{figure*}[thbp]
    \centering
    \includegraphics[width=0.8\textwidth]{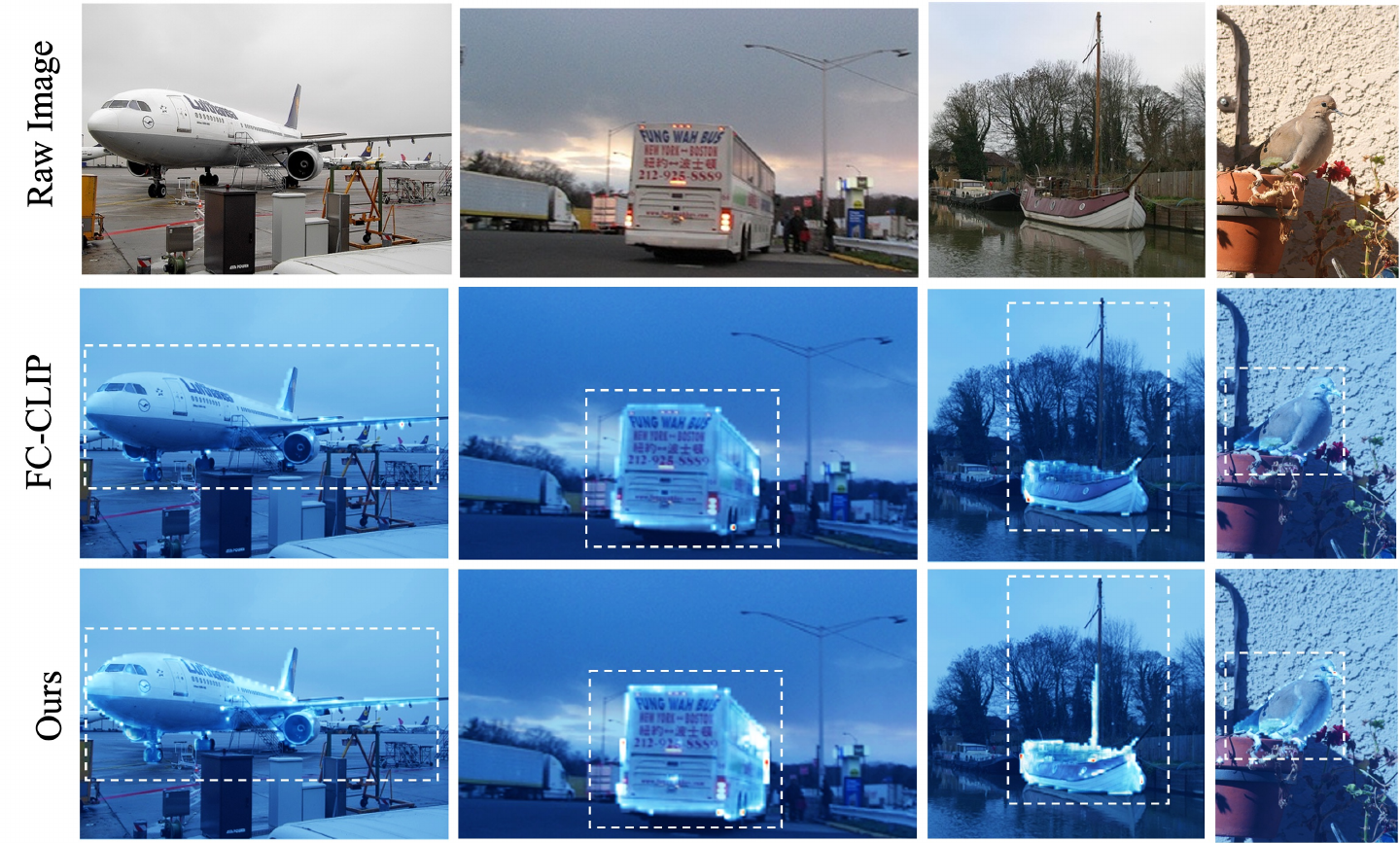}
    \caption{
    Comparative visualization of query attention maps on the PAS-21 dataset. The enclosed white box delineates the queried object intended for visualization. Our learned queries distinctly emphasize the object's boundaries and intra-content, showcasing the accuracy of attention allocation.}
    \label{fig:attention_map}
\end{figure*}
\begin{figure}[thbp]
\centering
    \includegraphics[width=0.98\linewidth]{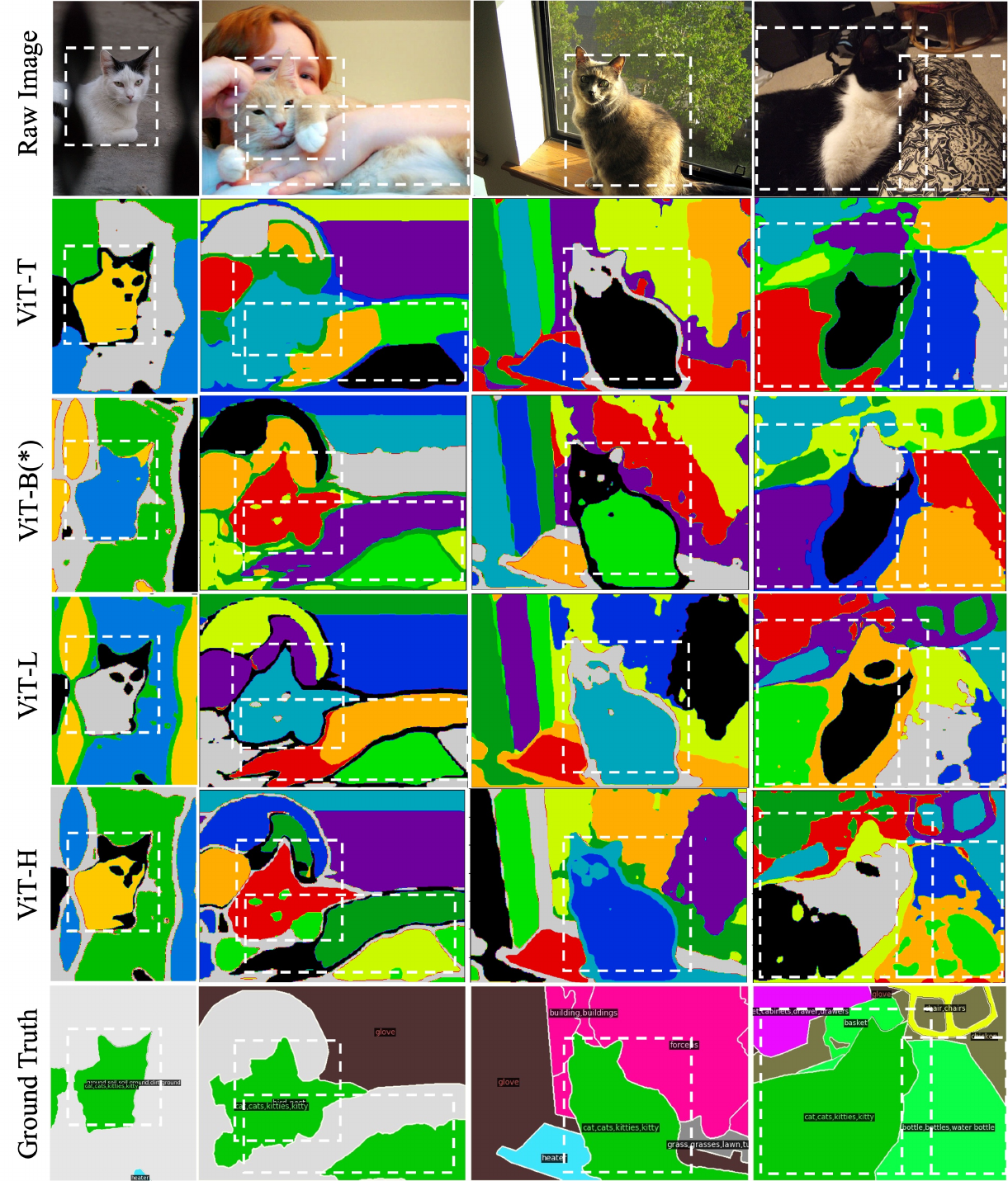}
    \caption{K-means clustering visualization of feature embeddings from various SAM image encoders on the PC-459 dataset. Note that the cluster colors are randomly assigned.}
    \label{fig:sam_variants}
\end{figure}

\section{Appendix}
Our supplementary material begins with the in-depth analysis of the FC-CLIP baseline's performance. Next, we explore more numerical results across various datasets, including evaluating mask recalls and an ablation study focusing on the co-training size of SAM. Finally, we present further qualitative visualization findings, featuring mask attention maps and segmentation results for two challenging datasets, A-847~\cite{ADE} and PC-459~\cite{pas_context}.

\subsection{Further discussion of FC-CLIP baseline}
FC-CLIP adopts a checkpoint selection strategy based on the PQ accuracy within the ADE20K benchmark~\cite{ADE}, a dataset known for its complexity with 150 diverse classes. Upon executing the FC-CLIP code and analyzing the final-round results, marked by * in Tab.~\ref{tab:fc-clip}, we observed tendencies towards overfitting and a subsequent decline in generalizability, This was accompanied by reduced effectiveness across various other open-vocabulary evaluation datasets, although there was improved performance on the COCO validation dataset.
Despite FC-CLIP's strategies to mitigate overfitting, the method's effectiveness in open-vocabulary scenarios, especially in the context of ADE20K's datasets, remains questionable. This raises concerns about the transparency of its model selection methodology. In contrast, our proposed framework, \shortname{}, which leverages the last iteration's checkpoints for inference, performs comparably on both the ADE20K and A-847 datasets. It demonstrates consistent and robust performance across all tested scenarios, thus eliminating the need for selective model evaluation.

\subsection{More numerical results }
\subsubsection{Comparative recall across datasets}
In Tab.~\ref{tab:recall_all}, we present the recall rates for our method and FC-CLIP across four datasets. This comparison is between the predicted mask proposals and class-agnostic semantic ground truth. We detail recall rates at IoU thresholds of 0.5, 0.75, and 0.9. The results demonstrate that \shortname{} generally outperforms in generalizing to mask proposals for unseen classes.

\subsubsection{Comparison with different SAMs}
In Tab.~\ref{tab:sam_all}, we provide detailed results of \shortname{} (w/o. mask ensemble), using ConvNeXt-L CLIP~\cite{clip} alongside different size of co-trained SAM~\cite{SAM}: ViT-T (Tiny)~\cite{mobile_sam}, ViT-B (Base), ViT-L (Large) and ViT-H (Huge). Across the board, the ViT-B configuration stands out, delivering better performance in our evaluations. We also provide visualizations of the k-means clustering results for feature embeddings from the SAM image encoders. The visualization demonstrates that ViT-B balances segmentation accuracy and connectivity, offering precise segmentation with good instance connectivity. ViT-T provides coarse boundaries, while ViT-L and ViT-H, though more precise, have reduced instance connectivity and may be less effective for panoptic segmentation with CLIP. Thus, ViT-B’s balanced performance makes it a robust choice.

\subsection{More qualitative visualizations}
\label{sec:sup-vis-res}

\subsubsection{Attention maps}
To illustrate the refinement of query features facilitated by injectors, we identify the query with the highest confidence and present its corresponding attention map from the final cross-attention layer within the transformer decoder. We map the attention map back to the original image for visualization purposes. The results are depicted in Fig. \ref{fig:attention_map}. It is evident that our queries exhibit heightened attention towards both the object boundaries and intra-content regions, indicative of the effectiveness of our approach in mask proposal generation.

\subsubsection{More results}
\begin{figure*}[th]
    \centering
    \includegraphics[width=0.98\textwidth]{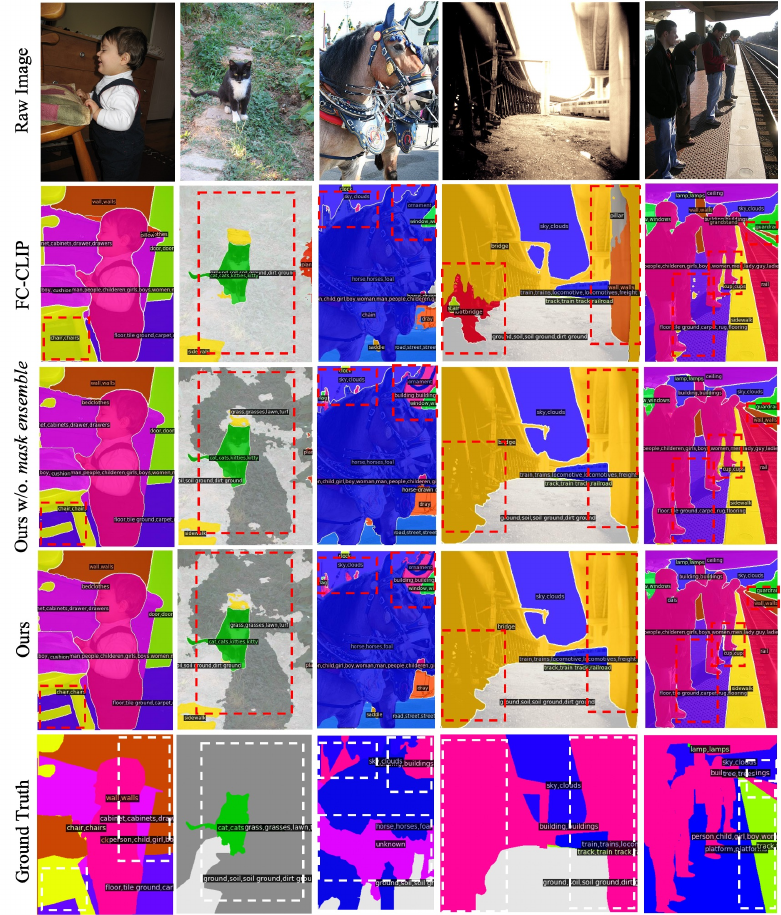}
    \caption{Qualitative visualizations in PC-459 dataset. Enhanced sections are delineated by red boxes, while white boxes underscore regions with imprecise ground truth annotations.}
    \label{fig:pc459}
    
\end{figure*}

\begin{figure*}[th]
    \centering
\includegraphics[width=0.98\textwidth]{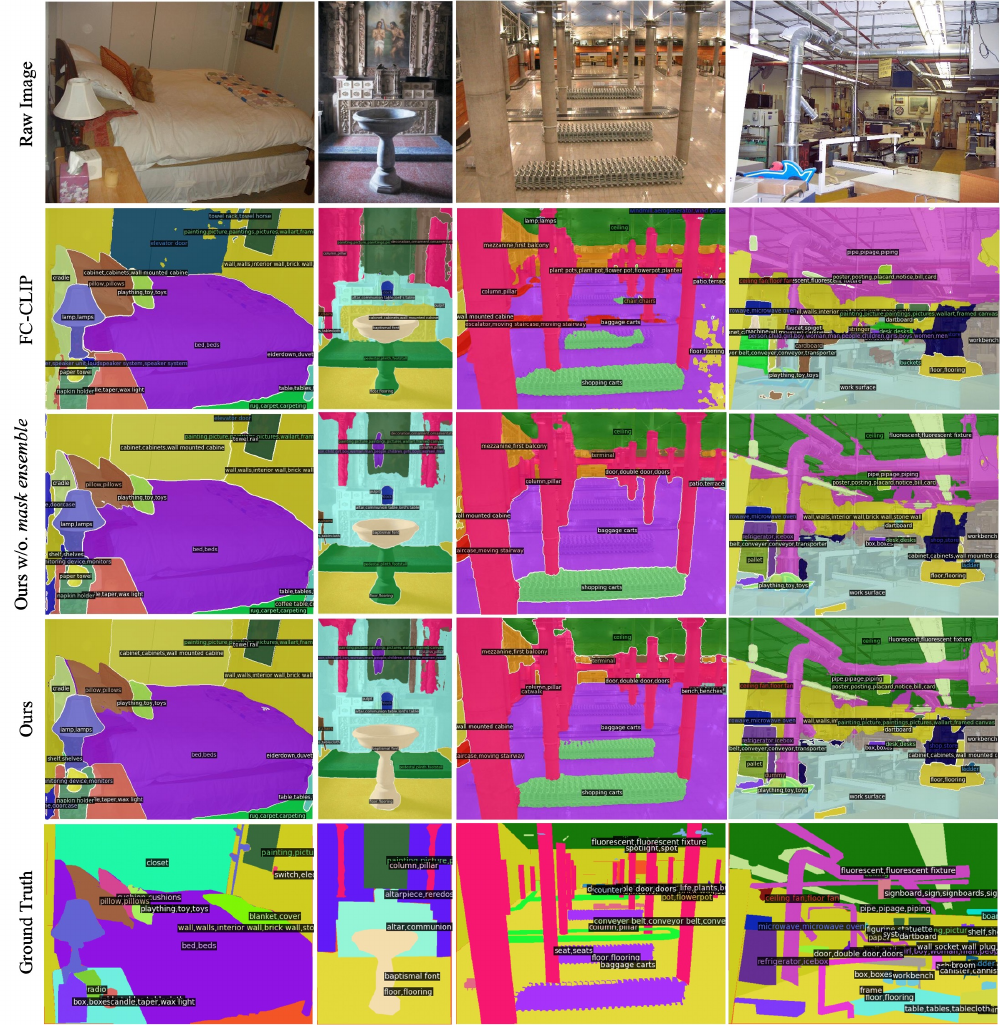}
    \caption{Qualitative visualizations in A-847 dataset.}
    \label{fig:a847}
\end{figure*}

We have expanded our visual comparisons in PC-459 dataset shown in Fig. \ref{fig:pc459}, and the A-847 dataset, depicted in Fig. \ref{fig:a847}. In Fig. \ref{fig:pc459}, it can be seen that our method generates more precise masks which are highlighted by red boxes.
Notably, we draw attention to the areas enclosed by white boxes, which exhibit coarse or imprecise annotations. For instance, the 'door' is overlooked in the first column, and the 'chair' annotations fail to precisely demarcate the chair legs. Meanwhile, in the second column, although the ground truth predominantly annotates the background as 'grass', a closer inspection reveals a composite of 'soil' and 'grass', with 'sidewalks' situated in the lower left quadrant. 
Fig. \ref{fig:a847} exemplifies the efficacy of our proposed method in producing high-quality masks, extending across a diverse array of novel classes. These include but are not limited to, 'toys', 'painted pictures', 'baptismal fonts', 'altars', 'decorative elements', 'columns', 'pipes', and 'fluorescent lighting', among others.


\clearpage

\onecolumn
\clearpage
\twocolumn
{\small
\bibliographystyle{ieee_fullname}
\bibliography{main}
}

\end{document}